\documentclass[11pt,runningheads]{llncs}

\usepackage{graphicx}
\usepackage{xcolor}
\usepackage{appendix}
\usepackage{subcaption}
\usepackage{url}
\usepackage{ amssymb }
\usepackage[font=small,labelfont=bf]{caption}

\newcommand{\Fol}{\mathcal{L}}

\newcommand\blfootnote[1]{
    \begingroup
    \renewcommand\thefootnote{}\footnote{#1}
    \addtocounter{footnote}{-1}
    \endgroup
}

\begin{document}
\title{IID Relaxation by Logical Expressivity: \\ 
A Research Agenda for 
Fitting Logics to Neurosymbolic Requirements
}
\titlerunning{IID Relaxation by Logical Expressivity}
%
\author{Maarten C. Stol~\inst{1}\thanks{Corresponding author: \texttt{maarten.stol@braincreators.com}}
\orcidID{0000-0002-5152-9188} 
\and \\
Alessandra Mileo~\inst{2}\orcidID{0000-0002-6614-6462} 
}
\authorrunning{Maarten C. Stol}
%
\institute{
BrainCreators, The Netherlands.
\\
\and
Insight Centre for Data Analytics at Dublin City University, Ireland
}

\maketitle              
%

%
\begin{abstract}
Neurosymbolic background knowledge and the expressivity required of its logic
can break Machine Learning assumptions about data Independence and Identical Distribution. 
In this position paper we propose to analyze IID relaxation in a hierarchy of logics 
that fit different use case requirements.
We discuss the benefits
of exploiting known data dependencies and distribution constraints for
Neurosymbolic use cases and argue that the expressivity required for this
knowledge has implications for the design of underlying ML routines.
This opens a new research agenda with general questions about
Neurosymbolic background knowledge and the expressivity
required of its logic.
\end{abstract}

%
%

%
\section{Introduction}
\label{sec:intro}

The relevance of Neurosymbolic (NeSy) methods 
\cite{besold2017neuralsymbolic,garcez2020neurosymbolicaithe,MARRA2024104062}
is apparent from the shortcomings 
of systems unable to combine learning with 
reasoning~\cite{marcus2020thenextdecade,thompson2020thecomputationallimits},
but fragmentation of the field remains an issue~\cite{DBLP:conf/nesy/OttLHH23}.
While important steps have been made 
by the categorization of common NeSy design patterns~\cite{bekkum2021modulardesignpatterns,mossakowski2022modulardesignpatterns}
a comparison of NeSy systems in terms of the logics they deploy is not available yet. 
This suggests the need for more formal approaches to compare and categorize NeSy formalisms 
by their use case requirements, 
for example, by a correspondence between required logical expressivity and known relations among Machine Learning (ML) data. 

Consider the central role of statistically Independent and Identically Distributed (IID) data~\cite{CaseBerg:01} in ML~\cite{bishop2007,prince2023understanding}, 
e.g., in supervised ML where all labeled data is \textit{assumed} to be drawn independently from the same joint distribution of labels and samples. 
We observe a striking contrast with the use of knowledge for supervision in Neurosymbolic uses cases, 
which are in some sense characterized by IID violations because
(1)~the expression of symbolic axioms breaks IID assumptions when the axioms relate samples or quantify out-of-distribution,  
and conversely
(2)~it is knowledge about true IID failures that is in turn expressed as relevant symbolic axioms. 

Comparisons on the logic side can be made by expressivity, i.e., the ability of a logic to separate classes of  
semantical structures on which its language is interpreted. 
Think of different flavours of OWL languages for Semantic Web~\cite{Horrocks_Patel-Schneider_McGuinness_Welty_2007}
and how their underlying Description Logics~\cite{conf/dlog/2003handbook,citeulike:550834}
and Modal Logics~\cite{BlackburnRijkeVenema01}
correspond to classes of structures fit for different use cases.
The expressivity and complexities of these logics can be studied in 
a hierarchy of First-Order Logic (FOL) fragments~\cite{10.1093/oso/9780192867964.003.0001}.

In this position paper
we propose an analogous approach to fit customized logical languages to the requirements of NeSy use cases 
\textit{by analyzing IID relaxation in a hierarchy of logics}. 
We seek to bridge the symbolic and sub-symbolic 
by the way logics for NeSy reasoning and supervision correspond to 
fundamental assumptions needed when learning from data. 
This opens a new research agenda on implications of IID relaxation, 
like the design of sample-dependency aware loss functions and batch selection. 
Moreover, an IID relaxation hierarchy may also shed light on related issues
like expressivity~vs.~scalability trade-offs in NeSy applications. 

%
%

%
\section{Motivation and Related Work}
\label{sec:motivation}

Assumptions of IID data have 
fundamental implications for ML, like 
maximum likelihood summing over independent samples, 
efficient gradient descent, 
cross validation making random splits, 
and deployment performance estimation from test data~\cite{prince2023understanding}. 
The success of applied ML may therefore seem remarkable given that IID assumptions typically fail in practice
\cite{rabanser2018failingloudlyan}.
From years of experience with commercial applications of ML we speculate 
the primary reason ML appears robust under IID violations is likely just \textit{use case selection bias}. 
In other words, use cases with more challenging requirements on sample dependencies or weaker distribution constraints 
are often not commercially successful and hence remain unreported. 
Similar issues of task selection bias exist in academic benchmark design~\cite{dehghani2021thebenchmarklottery}. 
This biased success despite pervasive IID violations 
may explain why non-IID ML remains relatively understudied.
Conversely, IID relaxation by robust NeSy methods can break out of this selection bias and potentially 
solve a much wider family of industrial problems. 

\smallskip
\noindent
\textbf{Benchmarks} ---
NeSy goals are closely tied to IID violation. 
Problems like MNIST addition~\cite{NEURIPS2018_dc5d637e} 
and Visual Sudoku~\cite{DBLP:conf/nesy/MorraABBCDGEGGK23}
pose classification of sample elements as a subtask for each sample, 
as do 
CLEVR~\cite{johnson2017clevradiagnostic} and 
Hand Written Formula~\cite{li2020closedloopneural}. 
These are interesting as NeSy benchmarks 
precisely because of the symbolic expression of their \textit{intra-}sample dependencies:
dependencies among sub-elements of a single sample. 
But lacking from these benchmarks and often present in the real world 
are \textit{inter-}sample dependencies: correlations \textit{between} instead of within samples.

In our experience industrial use cases typically do not suffer from a lack of available domain knowledge, 
making the use of symbolic axioms to compensate for data scarcity very attractive. 
The real challenge for AI in industry is therefore at the interface between scarce training data and 
an abundance of background knowledge. 
This asks for NeSy methods that also treat 
\textit{inter}-sample dependencies as first-class citizens at the sub-symbolic learning level,
as illustrated by the following use case. 


\vspace{-0.25cm}

\begin{figure}
    \centering
    \includegraphics[width=\textwidth]{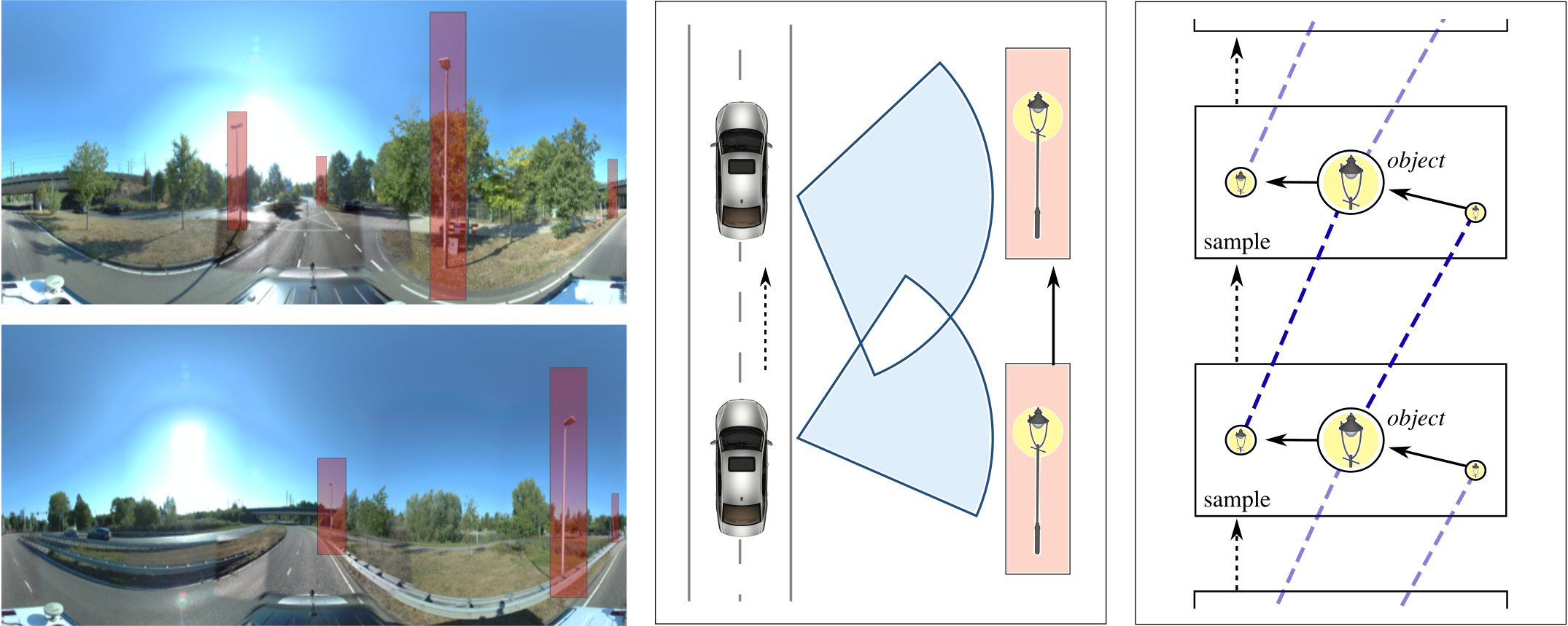}
    \caption{   Dataset for a streetlight detection use case,   
                a Knowledge Graph of \textit{inter}-, and \textit{intra-}sample relations. 
                \textbf{Left}: samples are consecutive images with bounding box object annotations in red. 
                \textbf{Middle}: top view diagram with overlapping samples taken at regular intervals. 
                \textbf{Right}: the resulting knowledge graph, 
                black arrows represent \textit{intra}-sample spatial relationships, 
                dotted arrows are \textit{inter-}sample relationships of sample contiguity, 
                dotted blue lines indicate shared membership of object equivalence classes.}
    \label{fig:streetlight}
\end{figure}

\vspace{-0.25cm}

\smallskip
\noindent
\textbf{Motivating use case} ---
For effective maintenance of streetlight systems the municipal databases require accurate information 
such as GPS, state of operation, and armature type. 
Regular observation by camera vehicles can be an attractive alternative to integration and maintenance of legacy database systems.
The ML object detection models typically
operate under IID assumptions~\cite{ZHANG2023120764}.
However, sample Independence evidently fails because it is known that 
(1) overlapping regions of two images should have correlated object detection probabilities, 
and (2) objects are spaced at regular intervals and unlikely to be in proximity. 
Training an object detection model without access to this knowledge  
may be data inefficient since known regularities are ignored and not explicit in its sample-based supervision signal. 
E.g.,~it cannot exploit known correlations between viewing angle and object appearance. 
As a result it may generalize poorly, unable to exploit object relationships 
to learn  representations that are robust for realistic variations of appearances. 

Figure~\ref{fig:streetlight} illustrates the dataset with \textit{intra}- and \textit{inter}-sample relationships. 
A logic would only have access to 
background axioms expressing object placement and other spatial regulatities in terms of 
node and object identities (constants), object class labels (unary predicates), 
and object relations (binary predicates). 
ML training can access vector data in the form of pixels or embeddings that remain hidden from the logic. 
Use cases like this emphasize the need and potential for NeSy methods in industry.

\textbf{Non-IID related work} ---
The understudied status of sample dependency in loss function design is demonstrated by its absence from otherwise excellent surveys 
on loss functions~\cite{ciampiconi2023asurveyand,wang2020acomprehensivesurvey}. 
But non-IID ML has been studied more generally;
some starters can be found in \cite{darrell_et_al:DagRep.5.4.18}. 
Augmenting feature vectors to capture hierarchical correlations between images appears in~\cite{10.5555/1625275.1625397}
and conditions for SVMs to perform well on non-IID data in~\cite{steinwart2009learningfromdependent}.
%
Graphs of spatiotemporal relations between images for better feature transformation appear in~\cite{Shi_Li_Gao_Cao_Shen_2017}.
Imputing missing regions in masked MNIST images is done 
by modeling \textit{intra}- and \textit{inter}-sample dependencies in~\cite{li2021partiallyobserved}. 
An index to quantify distribution shift between data splits based on 
metadata is experimentally shown to correlate with test error in~\cite{HE2021107383}. 
The same index appears in~\cite{casteels2020exploiting}
claiming correlations invariant for distribution shift are more likely to be causal and therefore robust to shift. 
An in depth treatment of
learning representations that are robutst under
interventions in the causal data generation process appears in~\cite{schölkopf2021towardscausalrepresentation}.

Learning from non-IID data is getting attention in the field of Federated Learning
e.g., see the discussion of clustering and categorization of skew types 
surveyed in~\cite{10.1016/j.neucom.2021.07.098}. 
Graph neural networks are essentially non-IID models~\cite{schlichtkrull2017modelingrelational} 
and federated learning on non-IID graphs appears in~\cite{tan2023federatedlearning}. 
Evaluating data skew based on deviations of accuracy scores among clients is presented in~\cite{Haller2023}. 
The Client Selection problem asks which clients to choose in each 
training round~\cite{10197174}.
While in Federated Learning this is primarily a problem of resource allocation and privacy, relevant for us is the fact that clients represent different training distributions. 
As such, heuristics for prioritizing clients based on their statistical utility might be useful in a 
Neurosymbolic environment 
to select one from many distributions that satisfy a given set of symbolic constraints.

%
%

%
\begin{center}
\begin{figure}[t]
\includegraphics[width=\textwidth]{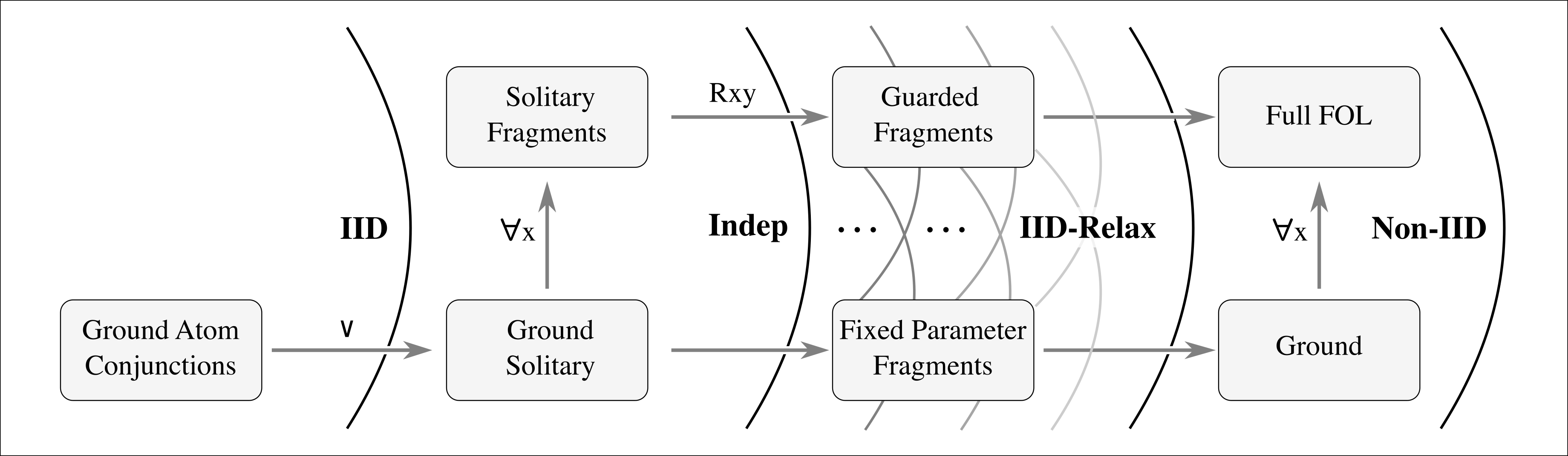}
\caption{\small A hierarchy of FOL fragments and their corresponding IID relaxations. 
The expressivity of ground atom conjunctions is sufficient for the 
symbolic content of standard (multi-label) ML datasets
but too weak to describe IID violations. 
Arrows indicate addition of logical
operators 
that result in incremental IID relaxations.  
The relevance of the Solitary Fragment, itself a Fixed Parameter fragment, 
is its ability to relax ID while maintaining full Independence. 
Intersections of 
Guarded- and Fixed Parameter languages
(indicated by overlapping regions) 
each capture a notion of IID relaxation. 
} 
\label{fig:Relaxation_Hierarchy}
\end{figure}
\end{center}

\vspace{-1.0cm}

\section{IID Relaxation}
\label{sec:relaxation}

Given a $d$-dimensional input space and $k$ different class labels,
ML models learn to map 
inputs $x\in\mathbb{R}^d$ to one-hot label vectors $y\in \{0,1\}^k$~\cite{bishop2007,prince2023understanding}.
Let $\Fol_G$ be the family of FOL fragments that contain only unary predicate symbols $P_y$ for each class label $y$, 
logical constants $c_x$ for each input $x$, and the propositional conjunction connective $\wedge$, 
i.e., $\Fol_G$ languages contain only conjunctions of positive\footnote
{
While relevant, a full treatment of negation falls outside the current scope.  
}   
ground atoms. 
Note that reasoning in $\Fol_G$ languages is logically trivial as every sentence is satisfiable. 
The interest in $\Fol_G$ is that it can express the symbolic content of ML (multi-class) labeled data,  
comparable to the treatment in~\cite{DBLP:conf/nesy/OttLHH23}.  

\textbf{Solitude} ---
Figure~\ref{fig:Relaxation_Hierarchy} shows the Ground Atom Conjunctions of $\Fol_G$ as the weakest FOL fragment of the hierarchy. 
We now extend languages in $\Fol_G$ with propositional disjunction $\vee$, but impose a condition called \textit{solitude}:
let Ground Solitary Fragments 
$\Fol_{G.Sol}$ 
be extensions of $\Fol_G$ with disjunction
\textit{such that conjuncts contain at most a single constant}, possibly more than once. 
Note that $\Fol_{G.Sol}$ languages maintain the ML Independence assumption
because they are too weak to express knowledge that conditions the classification of any sample on that of any other. 
On the other hand, this use of disjunction does break the assumption of Identical Distribution;
ID implies a unique joint distribution of $x$ and $y$, and $\Fol_{G.Sol}$ 
can express more general constraints on the set of all joint distributions by a disjunction of different predicates applied to the same constant. 

Moving up in Figure~\ref{fig:Relaxation_Hierarchy}, let 
$\Fol_{Sol}$ extend the languages in $\Fol_{G.Sol}$ with First-Order quantifiers and variables. 
An appropriate condition of solitude now demands that 
\textit{quantified subformulas contain no constants and at most a single variable}, possibly more than once. 
Note that by solitude, $\Fol_{Sol}$ languages again maintain the ML independence assumption
because they still can not express knowledge that conditions classification of any sample on that of any other. 
However, the language has now become sufficiently expressive for decidable taxonomy axioms, 
like concept inclusion axioms in Description Logics~\cite{conf/dlog/2003handbook,krötzsch2012adescriptionlogic}. 

Solitary Fragments exemplify 
Fixed Parameter Tractability:
keeping some parameter $p$ fixed to reduce the complexity of a class of problems~\cite{DBLP:series/faia/SamerS21}. 
Very broadly, Fixed Parameter Fragments (FPFs) in Figure~\ref{fig:Relaxation_Hierarchy} include all FOL fragments that can be defined in this way. 
Other approaches by which useful FPFs may be defined can be found 
in~\cite{gutin_et_al:DFU.Vol7.15301.179}.
Fixed-Parameter Tractability of logics appears 
in~\cite{FRICK20043}.

Next, including relation symbols $Rxy$ 
into the language 
brings Guarded Fragments (GF)~\cite{Andreka1998-ANDMLA-5} into range.
GFs are FOL fragments to which modal languages can be translated; 
they can be computationally well-behaved
while being strong enough to express many useful relational properties of graphs~\cite{lmcs:675}.
Various classes of structures of practical importance can be defined by modal axioms,
and GFs with constants~\cite{cate2005guardedfragmentswith} and counting quantifiers~\cite{pratthartmann2006complexity} exits in the literature. 

Examples formulae of all the language families discussed in this paper, from Ground Atom Conjunctions to Guarded- and Fixed Parameter fragments are provided in Appendix~\ref{app:examples}.

\medskip
\noindent
\textbf{IID Relaxation} ---
To the best of our knowledge 
this way of analyzing the relaxation of ML assumptions in terms of the expressive power of logical languages 
offers a novel perspective on Neurosymbolic research.
The rich literature about Guarded Fragments and Fixed Parameter Tractability
now acquires new relevance for the study of NeSy challenges. 
Moreover, we emphasize that this research doesn't necessarily proceed \textit{from} a given form of IID violation 
\textit{to} finding a corresponding FOL fragment. 
Instead, when a customized logic is designed to exactly fit the knowledge expressivity requirements 
of a NeSy use case 
\textit{that logic then \textit{captures} 
the corresponding notion of IID relaxation appropriate for that use case class}.
This in turn can drive the development of non-IID ML methods 
tailored for said notion of IID relaxation and all other use cases of that same class. 

Interesting candidates for this IID relaxation approach are captured by Fragments just outside IID in the overlap of GFs and FPFs. 
These fragments allow only weak expressions of inter-sample dependencies 
and can not force the joint distribution constraints very far from uniqueness. 
NeSy methods equipped with logics like this to learn from and reason with 
would be applicable to a range of new use cases, but with gradual and manageable increase of complexity. 
Think of bounded-horizon spatiotemporal use cases with small domain distribution shifts. 
If sample-dependency aware ML methods developed for these logics were shown empirically to outperform classical methods,
or even outperform existing NeSy methods
in their range of application, robustness, or data efficiency,
then the arguments of this position paper would be validated.
This in turn would lead to questions as to how far this research agenda can be pushed. 

Performance is, however, not the only motivation for this new research agenda. 
Aiming for a better understanding of NeSy challenges and methods is perhaps even more important. 
Hierarchies of logics ordered by language inclusion are also excellent tools for the 
comparison of the NeSy use cases to which they correspond and the methods aiming to solve them. 
Additionally, a language hierarchy as in Figure~\ref{fig:Relaxation_Hierarchy} is a mostly syntactical affair, 
but expressivity is concerned with interpreting a language on classes of model structures. 
Categorizing NeSy challenges along model theoretic properties of their underlying logics 
therefore provides yet another, more semantic lens through which we can analyse a use cases and the suitability of the applied NeSy formalism.

%
%

%
\section{Discussion and Conclusions}
\label{sec:consequences}

We now discuss consequences of these ideas and summarize our research agenda with a focus on 
dependency-aware loss functions and batch selection procedures, 
and the categorization of NeSy formalisms by means of the logical expressivity its background knowledge requires. 

\textbf{Loss calculation} --- 
ML loss functions for IID data like cross-entropy or maximum likelihood~\cite{prince2023understanding}
conveniently sum over individual data samples under assumption of Independence.  
Stochastic Gradient Descent~\cite{bottou2016optimizationmethods} 
applies the loss function to sample batches conveniently selected randomly under assumption of Identical Distribution. 
Any true dependencies or non-unique distributions that constitute IID violation 
will make these losses less representative of the true probabilities 
and hence negatively impact the learning task. 
It is therefore natural that loss function design for non-IID scenarios should 
be based on a logical analysis of background knowledge that takes 
sample dependencies and distribution constraints into account. 
In other words: 
\textit{known IID violations demand 
adequate and logically informed IID relaxation for the loss function and batch selection}. 

\textbf{Batch selection} --- 
A precondition for dependency-aware loss 
is that
batch loss should not depend on samples outside the batch. 
This can be achieved by characterizing
sample dependency relations in a Knowledge Graph (Figure~\ref{fig:streetlight}) 
and treating sample batches as subgraphs. 
A randomly selected batch can then be expanded along graph links to the smallest closure that contains it,
splitting disconnected components into separate batches. 
By invariance of modal satisfaction under generated submodels~\cite{BlackburnRijkeVenema01},
batches will preserve modal truths of the full dataset. 
This suggests batch selection followed by submodel generation can be a suitable approach 
to dependency-aware loss for modal FOL fragments. 

\textbf{Truth definitions} --- 
Following this trend 
we may consider directly translating modal truth definitions into NeSy loss terms.  
This does not appear in recent 
loss function surveys~\cite{ciampiconi2023asurveyand,wang2020acomprehensivesurvey}
but may be compared to Semantic Loss as proposed in~\cite{xu2018asemanticloss} 
and further developed in~\cite{ahmed2024semanticlossfunctionsneurosymbolic}.
Our aim would be to make sample loss a function of relational semantics on the Knowledge Graph of training data, 
analogous to modal truth defined by accessible states on the underlying modal frame~\cite{BlackburnRijkeVenema01}. 
NeSy use cases with sufficiently sparse dependency graphs, or for which loss calculation can be 
otherwise heuristically bounded 
can therefore point towards IID relaxations with tractable loss functions. 

\textbf{Model theory} --- 
More generally, fitting a logic to NeSy use case requirements will point to a set of 
model theoretic invariance results, and in turn 
to criteria for the design of NeSy batch selection procedures and loss functions. 
This would enable a comparison of NeSy formalisms along truly semantic lines of logical expressivity over 
classes of structures 
instead of our current syntactic approach by a hierarchy of language inclusions. 

\textbf{Conclusion} --- 
In this position paper we  propose a new research agenda to advance the idea that 
ties between formal knowledge and fundamental ML assumptions about  
training sample dependencies and distribution constraints
deserve to be first-class citizens in Neurosymbolic integration. 
Pursuing these ideas presents serious challenges both theoretical and for system design and can impact a range of other ML routines including calculation of
gradients, cross validation, estimation of train-test shifts, and searching over distributions.

\blfootnote{\textbf{Acknowledgements}: We extend our gratitude for time and resources provided by BrainCreators, 
for critical academic discussions with Erman Acar, Frank van Harmelen, Hanno Hildmann, and Rinke Hoekstra, 
and for work on the BrainCreators streetlight use case by Adem Günesen, Soroor Shekarizade, and Lucas Beerekamp. 
This research was supported by the 
Dutch Mkb-innovatiestimulering Regio en Topsectoren, 
Grant no. 823CFA64, and 
Science Foundation Ireland, Grant no. 12/RC/2289\_P2.}

\newpage

%
%

%
\appendix
\section*{Appendix}
%

%
\section{Example Formulae}
\label{app:examples}

The symbolic content of classical ML training data can be expressed in a language $\Fol_G$ of conjunctive ground atoms, 
with constants $x_i$ for input data, unary predicates ${P_y}_j$ for output class labels, and the $\wedge$ connective:
\begin{equation}
{P_y}_1(x_1) \wedge {P_y}_2(x_2) \wedge \ldots {P_y}_m(x_n) 
\end{equation}
Solitary Fragments $\Fol_{Sol}$ are defined as extensions of $\Fol_G$ with disjunction
such that 
\textit{conjuncts in} 
$\Fol_{Sol}$ CNFs  
\textit{contain at most a single constant} but possibly more than once.
This can express annotation ambivalence and multi-label annotations. 
\begin{equation}
({P_y}_1(x_1) \vee {P_y}_2(x_1))\, \wedge\, ({P_y}_3(x_2) \vee {P_y}_4(x_2))\, \wedge\, \ldots
\end{equation}
Predicates are loosely understood to represent label vectors and 
negation and implication are implicitly assumed. 
Hence, Taxonomy Axioms are universally quantified Solitary formulae and express label dependencies but not sample dependencies: 
\begin{equation}
\forall x 
\left[
({P_y}_1(x) \rightarrow {P_y}_2(x)) 
\right]
\end{equation}
Guarded Fragments require binary relation symbols and can express sample dependencies, 
either mixing with constants at ground level or as modal axioms about training data like spatiotemporal relationships. 
\begin{equation}
\forall x \left[\,
{P_y}_1(x) \rightarrow
\exists y (Rxy \wedge {P_y}_2(y))
\,\right]
\end{equation}
Propositional extension of $\Fol_{Sol}$ into Fixed-Parameter Fragments $\Fol_{FP}$ can be syntactical, 
like having at most 2 constants in each conjunct:
\begin{equation}
({P_y}_1(x_1) \vee {P_y}_2(x_1) \vee {P_y}_2(x_2))
\, \wedge\, 
({P_y}_3(x_3) \vee {P_y}_4(x_3) \vee {P_y}_4(x_4))
\end{equation}
And $\Fol_{FP}$ includes languages with structural constraints on formula graphs 
like the number of communities in a bipartite graph representation of the CNF.
An example combining syntactic and semantic constraints would be 2-CNF expressions 
with at most some fixed number of models $k$. 
Finally, quantified languages in $\Fol_{FP}$ extend these ideas with universal claims about label dependencies and 
training data structure. 
Overlap with Guarded Fragments is possible.

%
%

%
%
\bibliographystyle{splncs04}
\bibliography{mybibliography}

\begin{thebibliography}{10}
\providecommand{\url}[1]{\texttt{#1}}
\providecommand{\urlprefix}{URL }
\providecommand{\doi}[1]{https://doi.org/#1}

\bibitem{ahmed2024semanticlossfunctionsneurosymbolic}
Ahmed, K., Teso, S., Morettin, P., Liello, L.D., Ardino, P., Gobbi, J., Liang, Y., Wang, E., Chang, K.W., Passerini, A., den Broeck, G.V.: Semantic loss functions for neuro-symbolic structured prediction (2024)

\bibitem{Andreka1998-ANDMLA-5}
Andr\'{e}ka, H., N\'{e}meti, I., van Benthem, J.: Modal languages and bounded fragments of predicate logic. Journal of Philosophical Logic  \textbf{27}(3),  217--274 (1998). \doi{10.1023/a:1004275029985}

\bibitem{conf/dlog/2003handbook}
Baader, F., Calvanese, D., McGuinness, D.L., Nardi, D., Patel-Schneider, P.F. (eds.): The Description Logic Handbook: Theory, Implementation, and Applications. Cambridge University Press (2003), \url{http://dblp.uni-trier.de/db/conf/dlog/handbook2003.html}

\bibitem{bekkum2021modulardesignpatterns}
van Bekkum, M., de~Boer, M., van Harmelen, F., Meyer-Vitali, A., ten Teije, A.: Modular design patterns for hybrid learning and reasoning systems: a taxonomy, patterns and use cases (2021), \url{https://arxiv.org/abs/2102.11965}

\bibitem{besold2017neuralsymbolic}
Besold, T.R., d'Avila Garcez, A., Bader, S., Bowman, H., Domingos, P., Hitzler, P., Kuehnberger, K.U., Lamb, L.C., Lowd, D., Lima, P.M.V., de~Penning, L., Pinkas, G., Poon, H., Zaverucha, G.: Neural-symbolic learning and reasoning: A survey and interpretation (2017)

\bibitem{bishop2007}
Bishop, C.M.: Pattern Recognition and Machine Learning (Information Science and Statistics). Springer, 1 edn. (2007)

\bibitem{BlackburnRijkeVenema01}
Blackburn, P., de~Rijke, M., Venema, Y.: Modal Logic. No.~53 in Cambridge Tracts in Theoretical Computer Science, Cambridge University Press (2001)

\bibitem{bottou2016optimizationmethods}
Bottou, L., Curtis, F.E., Nocedal, J.: Optimization methods for large-scale machine learning (2016), \url{https://arxiv.org/abs/1606.04838}

\bibitem{lmcs:675}
Bárány, V., Gottlob, G., Otto, M.: {Querying the Guarded Fragment}. {Logical Methods in Computer Science}  \textbf{{Volume 10, Issue 2}} (May 2014), \url{https://lmcs.episciences.org/675}

\bibitem{CaseBerg:01}
Casella, G., Berger, R.: Statistical Inference. {Duxbury Resource Center} (June 2001)

\bibitem{casteels2020exploiting}
Casteels, W., Hellinckx, P.: Exploiting non-i.i.d. data towards more robust machine learning algorithms (2020)

\bibitem{cate2005guardedfragmentswith}
Cate, B., Franceschet, M.: Guarded fragments with constants (2005), \url{https://link.springer.com/article/10.1007/s10849-005-5787-x}

\bibitem{ciampiconi2023asurveyand}
Ciampiconi, L., Elwood, A., Leonardi, M., Mohamed, A., Rozza, A.: A survey and taxonomy of loss functions in machine learning (2023), \url{https://arxiv.org/abs/2301.05579}

\bibitem{darrell_et_al:DagRep.5.4.18}
Darrell, T., Kloft, M., Pontil, M., R\"{a}tsch, G., Rodner, E.: {Machine Learning with Interdependent and Non-identically Distributed Data (Dagstuhl Seminar 15152)}. Dagstuhl Reports  \textbf{5}(4),  18--55 (2015). \doi{10.4230/DagRep.5.4.18}, \url{https://drops-dev.dagstuhl.de/entities/document/10.4230/DagRep.5.4.18}

\bibitem{dehghani2021thebenchmarklottery}
Dehghani, M., Tay, Y., Gritsenko, A.A., Zhao, Z., Houlsby, N., Diaz, F., Metzler, D., Vinyals, O.: The benchmark lottery (2021)

\bibitem{10.5555/1625275.1625397}
Dundar, M., Krishnapuram, B., Bi, J., Rao, R.B.: Learning classifiers when the training data is not iid. In: Proceedings of the 20th International Joint Conference on Artifical Intelligence. p. 756–761. IJCAI'07, Morgan Kaufmann Publishers Inc., San Francisco, CA, USA (2007)

\bibitem{FRICK20043}
Frick, M., Grohe, M.: The complexity of first-order and monadic second-order logic revisited. Annals of Pure and Applied Logic  \textbf{130}(1),  3--31 (2004). \doi{https://doi.org/10.1016/j.apal.2004.01.007}

\bibitem{10197174}
Fu, L., Zhang, H., Gao, G., Zhang, M., Liu, X.: Client selection in federated learning: Principles, challenges, and opportunities. IEEE Internet of Things Journal  \textbf{10}(24),  21811--21819 (2023). \doi{10.1109/JIOT.2023.3299573}

\bibitem{garcez2020neurosymbolicaithe}
d'Avila Garcez, A., Lamb, L.C.: Neurosymbolic ai: The 3rd wave (2020), \url{https://arxiv.org/abs/2012.05876}

\bibitem{gutin_et_al:DFU.Vol7.15301.179}
Gutin, G., Yeo, A.: {Parameterized Constraint Satisfaction Problems: a Survey}. In: Krokhin, A., Zivny, S. (eds.) The Constraint Satisfaction Problem: Complexity and Approximability, Dagstuhl Follow-Ups, vol.~7, pp. 179--203. Schloss Dagstuhl -- Leibniz-Zentrum f{\"u}r Informatik, Dagstuhl, Germany (2017). \doi{10.4230/DFU.Vol7.15301.179}

\bibitem{Haller2023}
Haller, M., Lenz, C., Nachtigall, R., Awaysheh, F., Alawadi, S.: Handling non-iid data in federated learning: An experimental evaluation towards unified metrics (10 2023). \doi{10.1109/DASC/PiCom/CBDCom/Cy59711.2023.10361408}

\bibitem{HE2021107383}
He, Y., Shen, Z., Cui, P.: Towards non-i.i.d. image classification: A dataset and baselines. Pattern Recognition  \textbf{110},  107383 (2021)

\bibitem{Horrocks_Patel-Schneider_McGuinness_Welty_2007}
Horrocks, I., Patel-Schneider, P.F., McGuinness, D.L., Welty, C.A.: OWL: a Description-Logic-Based Ontology Language for the Semantic Web, p. 458–486. Cambridge University Press (2007)

\bibitem{citeulike:550834}
Hustadt, U., Schmidt, R.A., Georgieva, L.: A survey of decidable first-order fragments and description logics (2004), \url{http://citeseer.ist.psu.edu/hustadt04survey.html}

\bibitem{johnson2017clevradiagnostic}
Johnson, J., Hariharan, B., van~der Maaten, L., Fei-Fei, L., Zitnick, C.L., Girshick, R.: Clevr: A diagnostic dataset for compositional language and elementary visual reasoning (2017), \url{https://openaccess.thecvf.com/content_cvpr_2017/html/Johnson_CLEVR_A_Diagnostic_CVPR_2017_paper.html}

\bibitem{krötzsch2012adescriptionlogic}
Krötzsch, M., Simancik, F., Horrocks, I.: A description logic primer (2012), \url{https://arxiv.org/abs/1201.4089}

\bibitem{li2020closedloopneural}
Li, Q., , Huang, S., Hong, Y., Chen, Y., Wu, Y.N., Zhu, S.C.: Closed loop neural-symbolic learning via integrating neural perception, grammar parsing, and symbolic reasoning (2020), \url{http://proceedings.mlr.press/v119/li20f.html}

\bibitem{li2021partiallyobserved}
Li, Y., Oliva, J.B.: Partially observed exchangeable modeling (2021), \url{https://arxiv.org/abs/2102.06083}

\bibitem{NEURIPS2018_dc5d637e}
Manhaeve, R., Dumancic, S., Kimmig, A., Demeester, T., De~Raedt, L.: Deepproblog: Neural probabilistic logic programming. In: Bengio, S., Wallach, H., Larochelle, H., Grauman, K., Cesa-Bianchi, N., Garnett, R. (eds.) Advances in Neural Information Processing Systems. vol.~31. Curran Associates, Inc. (2018), \url{https://proceedings.neurips.cc/paper_files/paper/2018/file/dc5d637ed5e62c36ecb73b654b05ba2a-Paper.pdf}

\bibitem{marcus2020thenextdecade}
Marcus, G.: The next decade in ai: Four steps towards robust artificial intelligence (2020), \url{https://arxiv.org/abs/2002.06177}

\bibitem{MARRA2024104062}
Marra, G., Dumančić, S., Manhaeve, R., {De Raedt}, L.: From statistical relational to neurosymbolic artificial intelligence: A survey. Artificial Intelligence  \textbf{328},  104062 (2024). \doi{https://doi.org/10.1016/j.artint.2023.104062}, \url{https://www.sciencedirect.com/science/article/pii/S0004370223002084}

\bibitem{DBLP:conf/nesy/MorraABBCDGEGGK23}
Morra, L., Azzari, A., Bergamasco, L., Braga, M., Capogrosso, L., Delrio, F., Giacomo, G.D., Eiraudo, S., Ghione, G., Giudice, R., Koudounas, A., Piano, L., Cambrin, D.R., Risso, M., Rondina, M., Russo, A.S., Russo, M., Taioli, F., Vaiani, L., Vercellino, C.: Designing logic tensor networks for visual sudoku puzzle classification. In: d'Avila Garcez, A.S., Besold, T.R., Gori, M., Jim{\'{e}}nez{-}Ruiz, E. (eds.) Proceedings of the 17th International Workshop on Neural-Symbolic Learning and Reasoning, La Certosa di Pontignano, Siena, Italy, July 3-5, 2023. {CEUR} Workshop Proceedings, vol.~3432, pp. 223--232. CEUR-WS.org (2023), \url{https://ceur-ws.org/Vol-3432/paper19.pdf}

\bibitem{mossakowski2022modulardesignpatterns}
Mossakowski, T.: Modular design patterns for neural-symbolic integration: refinement and combination (2022), \url{https://arxiv.org/abs/2206.04724}

\bibitem{DBLP:conf/nesy/OttLHH23}
Ott, J., Ledaguenel, A., Hudelot, C., Hartwig, M.: How to think about benchmarking neurosymbolic ai? In: d'Avila Garcez, A.S., Besold, T.R., Gori, M., Jim{\'{e}}nez{-}Ruiz, E. (eds.) Proceedings of the 17th International Workshop on Neural-Symbolic Learning and Reasoning, La Certosa di Pontignano, Siena, Italy, July 3-5, 2023. {CEUR} Workshop Proceedings, vol.~3432, pp. 248--254. CEUR-WS.org (2023), \url{https://ceur-ws.org/Vol-3432/paper22.pdf}

\bibitem{pratthartmann2006complexity}
Pratt-Hartmann, I.: Complexity of the guarded two-variable fragment with counting quantifiers (2006)

\bibitem{10.1093/oso/9780192867964.003.0001}
Pratt-Hartmann, I.: {1Introduction}. In: {Fragments of First-Order Logic}. Oxford University Press (03 2023). \doi{10.1093/oso/9780192867964.003.0001}, \url{https://doi.org/10.1093/oso/9780192867964.003.0001}

\bibitem{prince2023understanding}
Prince, S.J.: Understanding Deep Learning. MIT Press (2023)

\bibitem{rabanser2018failingloudlyan}
Rabanser, S., Günnemann, S., Lipton, Z.C.: Failing loudly: An empirical study of methods for detecting dataset shift (2018), \url{https://arxiv.org/abs/1810.11953}

\bibitem{DBLP:series/faia/SamerS21}
Samer, M., Szeider, S.: Fixed-parameter tractability. In: Biere, A., Heule, M., van Maaren, H., Walsh, T. (eds.) Handbook of Satisfiability - Second Edition, Frontiers in Artificial Intelligence and Applications, vol.~336, pp. 693--736. {IOS} Press (2021). \doi{10.3233/FAIA201000}, \url{https://doi.org/10.3233/FAIA201000}

\bibitem{schlichtkrull2017modelingrelational}
Schlichtkrull, M., Kipf, T.N., Bloem, P., van~den Berg, R., Titov, I., Welling, M.: Modeling relational data with graph convolutional networks (2017), \url{https://arxiv.org/abs/1703.06103}

\bibitem{schölkopf2021towardscausalrepresentation}
Schölkopf, B., Locatello, F., Bauer, S., Ke, N.R., Kalchbrenner, N., Goyal, A., Bengio, Y.: Towards causal representation learning (2021), \url{https://arxiv.org/abs/2102.11107}

\bibitem{Shi_Li_Gao_Cao_Shen_2017}
Shi, Y., Li, W., Gao, Y., Cao, L., Shen, D.: Beyond iid: Learning to combine non-iid metrics for vision tasks. Proceedings of the AAAI Conference on Artificial Intelligence  \textbf{31}(1) (Feb 2017). \doi{10.1609/aaai.v31i1.10748}, \url{https://ojs.aaai.org/index.php/AAAI/article/view/10748}

\bibitem{steinwart2009learningfromdependent}
Steinwart, I., Hush, D., Scovel, C.: Learning from dependent observations (2009), \url{https://www.sciencedirect.com/science/article/pii/S0047259X08001097}

\bibitem{tan2023federatedlearning}
Tan, Y., Liu, Y., Long, G., Jiang, J., Lu, Q., Zhang, C.: Federated learning on non-iid graphs via structural knowledge sharing. (2023), \url{https://doi.org/10.1609/aaai.v37i8.26187}

\bibitem{thompson2020thecomputationallimits}
Thompson, N.C., Greenewald, K., Lee, K., Manso, G.F.: The computational limits of deep learning (2020), \url{https://arxiv.org/abs/2007.05558}

\bibitem{wang2020acomprehensivesurvey}
Wang, Q., Ma, Y., Zhao, K., Tian, Y.: A comprehensive survey of loss functions in machine learning (2020), \url{https://link.springer.com/article/10.1007/s40745-020-00253-5}

\bibitem{xu2018asemanticloss}
Xu, J., Zhang, Z., Friedman, T., Liang, Y., Broeck, G.: A semantic loss function for deep learning with symbolic knowledge (2018), \url{http://proceedings.mlr.press/v80/xu18h.html}

\bibitem{ZHANG2023120764}
Zhang, T., Dai, J., Song, W., Zhao, R., Zhang, B.: Oslpnet: A neural network model for street lamp post extraction from street view imagery. Expert Systems with Applications  \textbf{231},  120764 (2023). \doi{https://doi.org/10.1016/j.eswa.2023.120764}, \url{https://www.sciencedirect.com/science/article/pii/S0957417423012666}

\bibitem{10.1016/j.neucom.2021.07.098}
Zhu, H., Xu, J., Liu, S., Jin, Y.: Federated learning on non-iid data: A survey. Neurocomput.  \textbf{465}(C),  371–390 (nov 2021). \doi{10.1016/j.neucom.2021.07.098}, \url{https://doi.org/10.1016/j.neucom.2021.07.098}

\end{thebibliography}

\end{document}